\documentclass[conference]{IEEEtran}
\ifCLASSINFOpdf
\else
\fi

\usepackage{times}
\usepackage{soul}
\usepackage{url}
\usepackage[hidelinks]{hyperref}
\usepackage[utf8]{inputenc}
\usepackage[small]{caption}
\usepackage{graphicx}
\usepackage{amsmath}
\usepackage{amsthm}
\usepackage{booktabs}
\usepackage{algorithm}
\usepackage{algorithmic}
\usepackage{bm}
\usepackage{amsfonts}
\usepackage{multirow}
\usepackage{booktabs}
\usepackage{enumitem}
\usepackage{cite}
\usepackage{color}
\usepackage[switch]{lineno}
\usepackage{bm}

\begin{document}
%
\title{Heterogeneous Graph Pre-training Based Model for Secure and Efficient Prediction of Default Risk Propagation among Bond Issuers\thanks{$^{\star}$ Equal contribution. $^{\dagger}$ Corresponding author.}}

\author{

\IEEEauthorblockN{Xurui Li$^{\star}$$\dag$}
\IEEEauthorblockA{Fudan University\\
leexurui@gmail.com}
\and
 
\IEEEauthorblockN{Xin Shan$^{\star}$}
\IEEEauthorblockA{Bank of Shanghai\\
shanxin@bosc.cn}

\and
\IEEEauthorblockN{Wenhao Yin}
\IEEEauthorblockA{Shanghai Saic Finance Co., Ltd\\
yinwenhao@saicfinance.com}

\and
\IEEEauthorblockN{Haijiao Wang}
\IEEEauthorblockA{DD Global Inc.\\
wanghaijiaoleisure@gmail.com}

}


%


\IEEEoverridecommandlockouts
\makeatletter\def\@IEEEpubidpullup{3\baselineskip}\makeatother
\IEEEpubid{\parbox{\columnwidth}{
    {\fontsize{7.5}{7.5}\selectfont
    Accepted in NDSS Workshop 2024.}
}
\hspace{\columnsep}\makebox[\columnwidth]{}}

\maketitle

\begin{abstract}
Efficient prediction of default risk for bond-issuing enterprises is pivotal for maintaining stability and fostering growth in the bond market. Conventional methods usually rely solely on an enterprise's internal data for risk assessment. In contrast, graph-based techniques leverage interconnected corporate information to enhance default risk identification for targeted bond issuers. Traditional graph techniques such as label propagation algorithm or deepwalk fail to effectively integrate a enterprise's inherent attribute information with its topological network data. Additionally, due to data scarcity and security privacy concerns between enterprises, end-to-end graph neural network (GNN) algorithms may struggle in delivering satisfactory performance for target tasks. To address these challenges, we present a novel two-stage model. In the first stage, we employ an innovative Masked Autoencoders for Heterogeneous Graph (HGMAE) to pre-train on a vast enterprise knowledge graph. Subsequently, in the second stage, a specialized classifier model is trained to predict default risk propagation probabilities. The classifier leverages concatenated feature vectors derived from the pre-trained encoder with the enterprise's task-specific feature vectors. Through the two-stage training approach, our model not only boosts the importance of unique bond characteristics for specific default prediction tasks, but also securely and efficiently leverage the global information pre-trained from other enterprises. Experimental results demonstrate that our proposed model outperforms existing approaches in predicting default risk for bond issuers.
\end{abstract}


%

\section{Introduction}
Bond issuer default risk, a critical aspect of financial analysis, refers to the likelihood that the enterprise which issues a bond will become unable to fulfill its financial obligations, leading to a default~\cite{zhang2021study}. To make informed investment decisions and maintain market stability, it is crucial to comprehend and assess this risk. When assessing default risk for a bond issuer, traditional methods usually utilize its own operational data for prediction. However, real-world risk encompasses various factors such as the issuer’s financial health, market conditions, industry-specific risks, and overall economic stability, etc. There are often intricate interconnections among enterprises, leading to the spread of risks across the bond issuer landscape~\cite{zhang2023bond}. 




To effectively utilize information from associate enterprises to assist in assessing default risk for a target enterprise, a common approach is to employ graph-based risk propagation methods such as label propagation algorithm. However, these algorithms merely propagate known risk labels across the graph without effectively incorporating the inherent characteristics of each enterprise into the comprehensive prediction. To address this issue, one solution is to leverage graph embedding information besides the enterprises' own features when predicting risk propagation probabilities. While this approach merges the enterprises' inherent traits with their relational topology to some extent, it awkwardly combines these two types of information without fully capturing the dynamic propagation and evolution effects of the enterprises' features upon the entire corporate network structure. 

The recently popular graph neural network (GNN) methods such as GAT or Graphsage can well learn the diffusion pattern of node features in the graph~\cite{xiao2022graph}. However, there are several problems with directly using end-to-end GNN algorithms for our scenario. On the one hand, the performance of existing GNNs significantly suffers when distribution shifts occur between training and testing data, or when data scarcity is present. Bond issuers, typically large or even listed enterprises, possess a wealth of business information. However, other non-bond enterprises in the massive graph often lack such detailed features. GNN algorithms struggle to learn effective information from this kind of graph for training specific risk propagation tasks, because they cannot fully leverage the differences between bond issuers and other enterprises. On the other hand, in terms of security and privacy, it is not advisable for some information providers to disclose detailed operational information of related enterprises to business parties.

In this paper, we present a two-stage framework that utilizes graph pre-training techniques to predict the propagation probability of default risks among bond issuers. Initially, our model is pre-trained on a comprehensive enterprise knowledge graph (EKG). Subsequently, the embedding from the pre-trained encoder is combined with the unique information of the bond-issuing enterprise to train the default risk prediction model. The main contributions are summarized as follows:

\noindent$\bullet$ The proposed two-stage method can not only safely and effectively utilize the encrypted information from the massive EKG, but also fully leverage the abundant features of bond-issuing enterprises for risk prediction tasks.

\noindent$\bullet$ We also introduce a novel Masked Autoencoders for Heterogeneous Graph (HGMAE) model to learn a better representation during the pre-training of heterogeneous graphs.

\noindent$\bullet$ Extensive experiments show that our method outperforms the state-of-the-art baselines in the risk dataset, indicating the effectiveness of the two-stage framework, as well as the HGMAE model.

\begin{figure*}
    \centering
    \includegraphics[width=\textwidth]{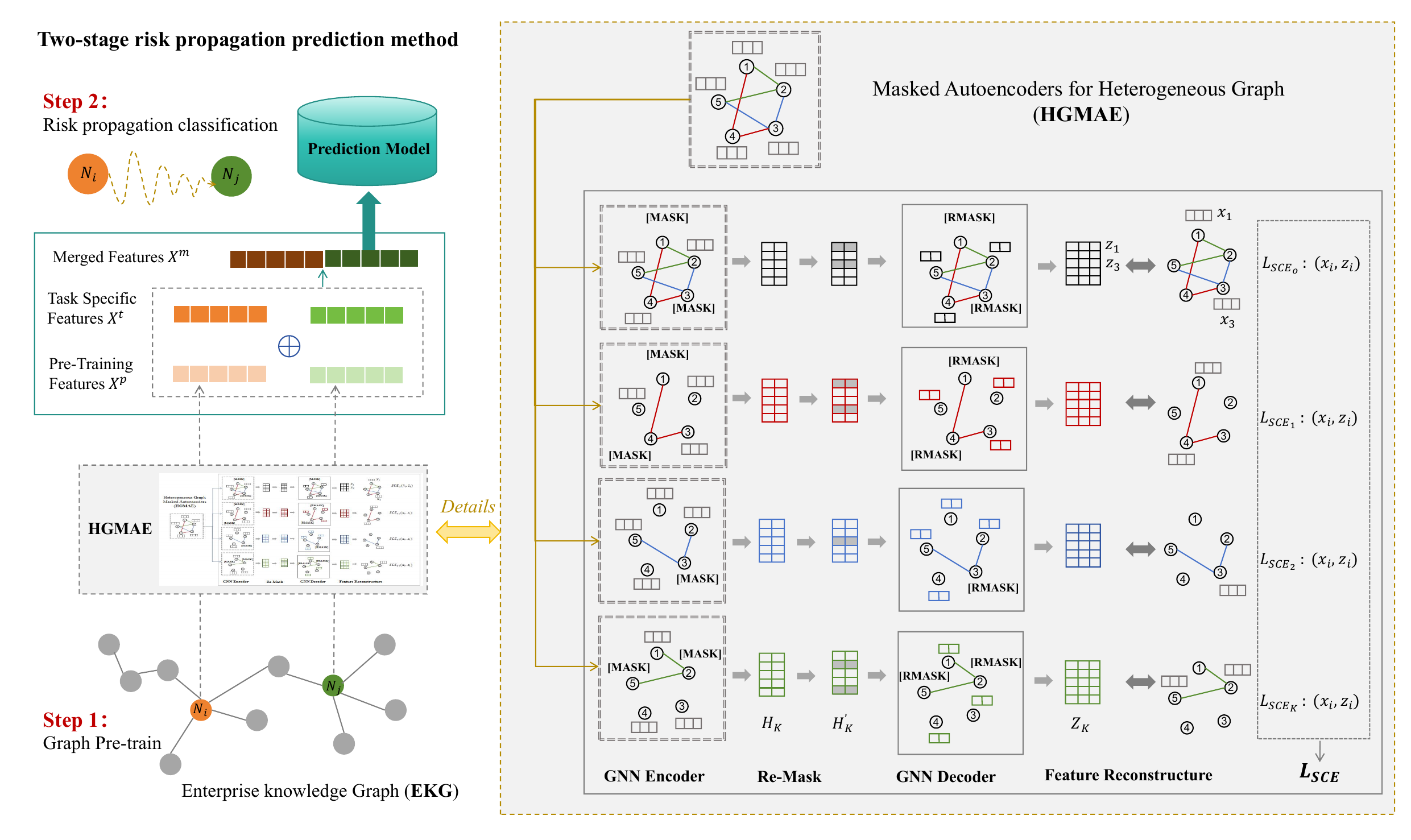}
    \caption{The main framework of our two-stage risk propagation prediction method. The right part shows the details for the proposed HGMAE model. Different Colors for HGMAE indicate different edge types. }
    \label{fig: mainModel}
\end{figure*}

\section{Related Works}
\vspace{3pt}\noindent\textbf{Pre-trained Graph Embeddings}. The first-generation pre-trained graph models aim to produce effective graph embeddings for diverse tasks~\cite{xia2022survey}. DeepWalk, a pioneer in this field, introduced the concept of graph embedding by treating the paths traversed by random walks over graphs as sentences. It utilizes skip-gram to learn latent node representations. Following in DeepWalk's footsteps, Node2vec developed a flexible approach to define a node's network neighborhood and designed a biased random walk procedure to efficiently explore diverse neighborhoods. Furthermore, several researchers have also attempted to learn embeddings for heterogeneous graphs, sub-graphs, and molecular graphs, such as sub2vec~\cite{cai2018comprehensive}, sub-graph2vec, etc.

\vspace{3pt}\noindent\textbf{Pre-trained Graph Encoders}. With the emergence of expressive GNNs and Transformer, recent methods have embraced a transfer learning setting where the goal is to pre-train a generic encoder that can deal with different tasks. Compared to the pre-trained graph embedding methods, pre-trained graph encoders can provide a better model initialization, which usually leads to a better generalization performance and speeds up convergence on the target tasks. In addition, the modern models are usually trained with larger scale database, more powerful or deeper architectures, and new pre-training tasks~\cite{xia2022survey, lu2021learning}. Their architectures broadly fall into two categories: graph neural networks (e.g., GIN, HAN)~\cite{GIN} and hybrid of GNNs and Transformer (e.g., MPG, HGT)~\cite{HGT}. Their pre-training strategies can be widely divided into supervised and unsupervised ones. Although the supervised pre-training brings remarkable improvements, they often require domain-specific knowledge which significantly limits their wider applications. More importantly, some supervised pre-training tasks might be unrelated to the downstream task of interest and can even hurt the downstream performance. As for unsupervised graph pre-training, it can be mainly divided into four categories: 1) Graph autoencoders: typical models such as GAE, VGAE, SIGVAE use self-supervised graph reconstruction for learning discriminative representations~\cite{SIGVAE}. 2) Graph autoregressive modeling: typical models such as GPT-GNN, MGSSL perform the autoregressive reconstruction on given graphs iteratively instead of reconstruct the graph all at once~\cite{gptgnn}. 3) Masked components modeling: typical model such as GROVER~\cite{GROVER} masks out some components from the graph and then trains the model to predict them. 4) Graph contrastive learning: typical models such as InfoGraph~\cite{Infograph}, GMI~\cite{GMI} use deep infoMax method for node and graph-level representation learning, while other models such as SimGRACE~\cite{SimGRACE}, GraphCL~\cite{GraphCL} and CCA-SSG~\cite{CCASSG} use instance discrimination method for contrastive learning.


\section{Methodology}

Our objective is to predict the defaulting probability of a target bond issuer enterprise when its associated source bond issuer defaults. To achieve this goal, we have employed the following strategies.

\vspace{3pt}\noindent\textbf{Construct Enterprise Knowledge Graph}. To begin, we develop an extensive enterprise knowledge graph (EKG) as a foundation for graph pre-training~\cite{li2023stinmatch}. This EKG connects various enterprises through diverse edge types, including ``\textit{parent-subsidiary}'', ``\textit{share-investor}'', ``\textit{share-manager}'', ``\textit{share-legal-person}'' and ``\textit{invest-by}'', etc. Each enterprise node within this graph is characterized by a set of common properties. These properties range from fundamental business information (such as registered capital and number of employees) to basic operating information like net profit and net income. Furthermore, historical risk features are also involved (such as the number of administrative penalties received and the number of litigations).
 
\vspace{3pt}\noindent\textbf{Graph Pre-training}. Subsequently, we proceed to implement graph pre-training on the meticulously constructed EKG. Following the approach of GraphMAE~\cite{hou2022graphmae}, we adopt feature reconstruction as the core training objective. Specifically, we propose a HGMAE model for heterogeneous graph pre-training. This method entails masking certain features within the graph and then challenging the model to accurately predict and reconstruct these masked attributes, thereby enhancing its understanding and representation capabilities. The backbones of the encoder and decoder for HGMAE can be any type of GNNs and we use GAT~\cite{xiao2022graph} here.

Formally, we implement a procedure where we select a subset of nodes and apply a masking technique by replacing each of their features with the designated mask token [MASK]. The core objective is to accurately reconstruct the masked node features based on the partially observed node signals and the provided input adjacency matrix. We use the scaled cosine error (SCE) for measuring the reconstruction performance. Different from GraphMAE, our model not only implement the feature reconstruction on the whole graph, but also involves the reconstruction for each isomorphic subgraph. As shown in Fig.~\ref{fig: mainModel}, different edge types are shown in different colors. The first workflow of calculating the reconstruction error $\mathcal{L}_{SCE_{o}}$ for the original graph is the same to that of GraphMAE. For each edge type $e_{k}$, we obtain its corresponding subgraph $G_{k} = (V_{k}, A_{k}, X_{k})$ where $V_{k}$ is the vertex collection, $A_{k}$ is the adjacent matrix and $X_{k}$ is the feature vector. Then we sample a subset of nodes $\hat V_{k} \in V_{k}$ and mask each of their features with a mask token [MASK].

To identify the nodes for masking, we employ a uniform random sampling technique which is instrumental in maintaining an unbiased enhancement and recovery of features within GNNs because each node inherently draws upon the attributes of its neighboring nodes. In order to establish a challenging self-supervised learning task that generates significant and resilient node representations, we have set the masking ratio at 50\%. However, the utilization of the [MASK] token can introduce a potential inconsistency between the training and inference stages because this token does not naturally appear during the inference process. To alleviate this issue, rather than consistently replacing masked features with the explicit [MASK] token, we have introduced an additional strategy: there is a 15\% probability that a masked feature will be substituted with a randomly chosen token. This strategic variation serves to cultivate more diverse and superior-quality node representations, which in turn boosts the overall performance of the model. To enhance the encoder's learning of compressed representations, we replace the latent vector $H_{k}$ on masked node indices with another mask token [RMASK] into $H^{'}_{k}$. This approach enables the GNN decoder to recover input features for a node based on a group of nodes rather than just the node itself. This, in turn, aids in teaching the encoder to generate high-level latent code. Given the original feature $X_{k}$ and reconstructed output $Z_{k}$ from the decoder, we define the reconstruction error for the $k_{th}$ subgraph as:
\begin{equation}
    \mathcal{L}_{SCE_{k}} = \frac{1}{\hat V_{k}} \sum_{v_{ki} \in \hat V_{k}}(1-\frac{x^{T}_{ki}z_{ki}}{\|x_{ki}\|.\|z_{ki}\|})^{\gamma}
\end{equation}

Finally, we merge all the reconstruction errors from original and all $K$ subgraphs into the entire loss:
\begin{equation}
    \mathcal{L}_{SCE} = \mathcal{L}_{SCE_{o}} + \frac{\eta}{K} \sum^{K}_{k=1} \mathcal{L}_{SCE_{k}}
\end{equation}

\noindent Here $\gamma$ and $\eta$ are hyper-parameters and are all set to 1 for our experiments. For downstream applications, the encoder is applied to the input graph without any masking in the inference stage. The generated node embeddings can be used for various graph learning tasks, such as node classification and graph classification. 

\vspace{3pt}\noindent\textbf{Default risk propagation prediction}. We build the task-related propagation pairs following the corresponding construction steps:
1) Taking the bond issuers who have defaulted as the black seed nodes.
2) Expand each seed node outward $N$ rounds through EKG (here we set $N$ to 3), and combine the target nodes (which also belongs to bond issuers) with the seed node to form propagation pairs.
3) For these pair samples, we label the pair as a black sample only when the target enterprise defaults after the seed node, and the rest are white samples.
4) Randomly select a certain number of white sample pairs for balance.

For each enterprise node in the pair, we concentrate its task-specific feature vector $X^{t}$ with the output graph embedding vector  $X^{p}$ from HGMAE as the fusion vector $X^{m}$. We combine the fusion vectors $X^{m}$ for the source and target enterprises as the final merge vector for training a supervised classification model. This model can be used to predict the default risk propagation property between bond issuers.
 
\section{Experiments}
\vspace{3pt}\noindent\textbf{Dataset}. we build the dataset for risk propagation prediction task using real bond-issuing market information of China. Firstly, we build the EKG as described in the methodology section. A total of 6714 bond issue entities are used as seed enterprises for EKG. After 5 round of expansion using different relationships, we form the EGK of more than 20 million nodes. The input dimension of the basic features for graph pre-training is 124. The output embedding size of the graph pre-training model is set to 256. We obtain 3658 propagation pair samples, with half of them are black. We divide 80\% of them as training set and 20\% as test set.


\begin{table}
  \centering
  \begin{tabular}{ c| c| c }
  \toprule
Pretrain Method & Base Model & Micro-F1 \\
 \midrule
None & Logisti & 0.764 \\
None & Random Forest & 0.785 \\
None & GBDT & 0.791 \\
None & XGBoost & 0.794 \\
None & GCN & 0.792 \\
None & GAT & 0.806 \\
None & GraphSage & 0.813 \\
DeepWalk & XGBoost & 0.798 \\
Node2Vec & XGBoost & 0.801 \\
GAE & XGBoost & 0.811 \\
GPT-GNN & XGBoost & 0.816 \\
GRACE & XGBoost & 0.814 \\
InfoGCL & XGBoost & 0.822 \\
GraphMAE & XGBoost & 0.825 \\
\textbf{HGMAE} & \textbf{XGBoost} & \textbf{0.831} \\
 \bottomrule
\end{tabular}
\caption{Experimental results for different methods on the risk dataset.}
\label{tab:performance}
\end{table}

\vspace{3pt}\noindent\textbf{Result discussion}.
The detailed comparisons between our method and other baselines are shown in Table~\ref{tab:performance}. It can be found that XGBoost performances the best among the traditional classification methods. Directly using GNN models as classifiers does not perform very well. We then use the XGBoost as the base classifier for the two-stage method. Compared to solutions that only rely on the basic features of the enterprise itself, simple graph embedding methods such as Deepwalk and Node2vec can improve the performance to a certain extent. Graph embeddings output from the pre-trained graph models such as GAE and GraphMAE perform better than that of simple graph embedding. It is because they involve both node features and graph structures during the pre-training. Our method HGMAE outperforms all other methods in the two-stage method. Compared to the previous state-of-the-art method GraphMAE, HGMAE not only calculate the reconstruction error for the original graph, but also implement the mask reconstruction on each separate isomorphic subgraphs. The importance and data distribution of different relationship types are not consistent. If we  indiscriminately blend all edges for the purpose of learning, there is a potential that critical information encapsulated in significant edges with a sparse count would be easily attenuated or overshadowed by the sheer volume of other less important relationship-type edges. HGMAE employs a distinctive approach by categorically segregating different edge types and individually handling the reconstruction errors associated with each of them. It helps learn the feature diffusion better and improves the model performance.

\section{Conclusion}
In this study, we introduce a two-stage framework for predicting the default risk propagation among bond issuers. In the initial phase, we utilize the novel HGMAE model to implement heterogeneous graph pre-training on the extensive enterprise knowledge graph. This approach ensures safe and efficient learning and utilization of the general node representation from a global perspective. Subsequently, in the second stage, we integrate task-specific features with the pre-trained embeddings of both the source and target enterprises to train the risk propagation prediction model. Extensive experimental results demonstrate the efficiency of our two-stage framework, as well as the HGMAE model.






%

\end{document}